# AUTOMATIC LOCATION DETECTION BASED ON DEEP LEARNING

Anjali Karangiya [1], Anirudh Sharma [2], Divax Shah[3], Kartavya Badgujar [4], Dr. Chintan Thacker [5], Dainik Dave [6].
*Department of Computer Science and Engineering*
*Parul University*
Vadodara, India.
[1]anjalikarangiya@gmail.com,[2]sharmaanirudh545@gmail.com, [3]divax12345@gmail.com, [4]kartavyabadgujar@gmail.com,
[5]chintanthacker450@gmail.com, [6]dainikdave870@gmail.com.

*Abstract*—The proliferation of digital images and the advancements in deep learning have paved the way for innovative solutions in various domains, especially in the field of image classification. Our project presents an in-depth study and implementation of an image classification system specifically tailored to identify and classify images of Indian cities. Drawing from an extensive dataset, our model classifies images into five major Indian cities: Ahmedabad, Delhi, Kerala, Kolkata, and Mumbai to recognize the distinct features and characteristics of each city/state. To achieve high precision and recall rates, we adopted two approaches. The first, a vanilla Convolutional Neural Network (CNN) and then we explored the power of transfer learning by leveraging the VGG16 model. The vanilla CNN achieved commendable accuracy and the VGG16 model achieved a test accuracy of 63.6%. Evaluations highlighted the strengths and potential areas of improvement, positioning our model as not only competitive but also scalable for broader applications. With an emphasis on open-source ethos, our work aims to contribute to the community, encouraging further development and diverse applications. Our findings demonstrate the potential applications in tourism, urban planning, and even real-time location identification systems, among others.

## I. INTRODUCTION

In most recent years, the advent of digital photography and ubiquity of smart devices have to led to a massive surge in the volume of digital images being produced and shared by daily Coupled with the progress in computational capabilities and the development of deep learning techniques, the realm of image processing has opened up new horizons, especially in Image classification [5], a subset of computer vision [10,15], is the process of categorizing images into one of several classes or labels.

### A. Background

In the modern era, where data takes center stage, images stand as one of the most pervasive forms of data available. They tell stories, hold histories, and convey vast amounts of information in a singular frame. One particular subset of image data that holds a wealth of untapped potential is city imagery. As the world becomes increasingly urbanized, understanding, classifying, and leveraging images of cities is of paramount importance. From the avenues of Mumbai to the streets of Delhi, each city has its unique fingerprint in the form of architecture, street layouts, and landmarks.

### B. Importance of Image Classification

The importance of image classification in computer vision and artificial intelligence is well-established[21]. It's the stepping stone to more complex tasks like object detection, image segmentation, and scene recognition. However, its application in classifying and recognizing distinct cities holds unique challenges and opportunities. Recognizing a city from an image isn't merely about identifying buildings or streets; it's about capturing the essence, the culture, and the unique blend of modernity and history that each city holds.

### C. Objective of the research

Given the profound potential of city image classification, this research aims to bridge the gap between generic image classifiers and a specialized tool capable of recognizing Indian cities with high accuracy. We endeavor to explore the capabilities of CNNs from scratch and harness the power of pre-trained models, navigating the spectrum from Vanilla CNNs to the intricacies of the VGG16 [6] model with fine-tuning. Through this journey, we aim to contribute a robust model to the community, fostering further exploration and real-world applications in this niche yet significant domain.

## II. RELATED WORK

This paper describes the provision of location-based services using non-GPS image acquisition. The system is designed to help city visitors access more information via mobile phones and built-in cameras. GPS has become a popular navigation tool. GPS data provides high accuracy, but it is not always present on all handheld devices and is not always precise in all situations, such as in urban areas[1]. The emergence of huge size of geo-calibrated image data is an important reason why computer vision is starting to look at the global scale[15]. Image geolocalization is the difficult task of estimating the location of an image based on its visual content in the form of GPS coordinates[2]. Photographic geolocation is a difficult task because most photographs present multiple, sometimes vague, unknown locations[7]. We can use geotagging of labeled images to segment cities into locations of interest, use a training sequence to record priors and transition probabilities between those locations, and extract visual features from each image[14]. Geospatial image tagging has many applications, including image retrieval, image editing, and scanning[12]. Landmark recognition is a complex task in the field of image classification due to the numerous architectural solutions of different landmarks[6]. Since the images of these characters are often diverse, listing the test data is difficult due to many differences in appearance and many outliers[18]. The purpose of this task is to automatically generate many images for a given object class. A multimodal approach using text, metadata, and visual features is used to collect many high-quality images from the web[19]. The recent advent of large-scale image processing makes data association (such as brute force scene matching) very efficient[15]. It proposes an automatic algorithm that collects data from the internet and collects hundreds of images for a group of questions[19]. When querying the database for images from different locations, the percentage of images with low similarity is usually greater than the percentage of images with high similarity. So if your query scene is in the database you get a higher response[1]. For object detection in early time methods used shape-based matching, but generally did not cope with solution of this problem is proposed like use of clustering, greedy N-optimal path matching, ratio tests, and Hough transforms methods[3]. For efficient land recognition, RMSprop showed the best results compared to other optimizers for tuning the model[6]. Determining the location of the building is based on four specific images: average color, most frequent colors, line energy, and tiny images[8]. Various evaluation standards can be used for the location accuracy, together with the average positioning errors, positioning standard deviation, and circle probability error[9]. Convolutional neural networks (CNNs) have been very successful in computer vision, but their inability to incorporate anatomical space into the decision process has hindered the success of some form of image analysis[20]. Convolutional neural networks are deep learning algorithms that can train large data sets using millions of 2D image parameters as input and convolve them with filters to produce the desired result[10]. Despite recent advances in deep learning, computer vision for building search can be overwhelming in many areas and locations. Modeling to detect specific buildings in global satellite images is even more difficult[13]. Image quality is the focus of Google Lens. Therefore, it is necessary to classify each image according to its main quality characteristics: composition, resolution and digital noise[4].

### III. PROPOSED WORK AND METHODOLOGY

Our dataset comprises images from five major Indian cities: Ahmedabad, Delhi, Kerala, Kolkata, and Mumbai. These images were sourced from various online platforms, ensuring a rich mix of visuals – from skyline shots to street-level captures. Given the varied sources of our images, it was essential to maintain consistency across the dataset. The preprocessing involved:

  - Resizing: All images were resized to a consistent dimension of 175x175 pixels.
  - Normalization: Image pixel values were scaled to fall between 0 and 1, ensuring numerical stability during model training.

  - Data Splitting: The dataset was divided into training set (70%), validation set (15%), and test (15%) set. This division aids in model evaluation and prevents overfitting[1].

The core of our research was built on a stepwise exploration of different model architectures. From starting with a straightforward Vanilla CNN to harnessing the power of an established model like VGG16, our methodology was designed to test, iterate, and refine.

#### A. Vanilla CNN model

**1. Architecture**
Our initial model, referred to as the Vanilla CNN, was a foundational approach to image classification. The architecture consisted of:
  - Multiple convolutional layers for feature extraction, each followed by a batch normalization layer to stabilize activations.
  - Max pooling layers for spatial down sampling.
  - Dropout layers introduced at various stages for regularization and to mitigate overfitting[6].
  - Fully connected dense layers towards the end, culminating in a softmax layer for classification.

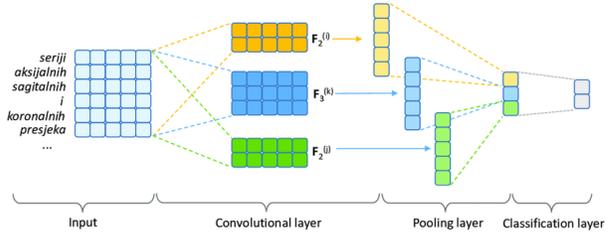

**Figure 3.1: Vanilla CNN Architecture [22]**

**2. Training Parameters**
The model was trained using the Adam optimizer with a categorical cross-entropy loss function, which is standard for many classification functions. Additionally, callbacks like early stopping and learning rate reduction were incorporated to optimize the training process.

#### B. Transfer Learning with VGG16

**1. Introduction to VGG16**
VGG16 is a deep convolutional network designed by the Visual Geometry Group at Oxford, known for its simplicity and high performance on the ImageNet dataset[7]. Given its success in generic image classification, it presented itself as an ideal candidate for transfer learning.

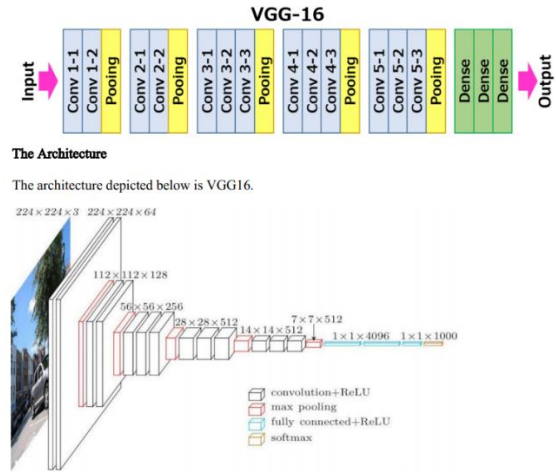

**Figure 3.2 VGG16 Architecture [23]**

**2. Integration with our project**
Incorporating VGG16, the initial layers were frozen, allowing us to utilize its already trained weights. Our dataset's images were then passed through VGG16, and the output was fed into custom dense layers tailored for our specific classification task.

#### C. Fine-Tuning VGG16

**1. Need for fine-tuning**
While VGG16 offered a robust start, its pre-trained weights were based on a different dataset (ImageNet). To make it more attuned to our city images, a fine-tuning process was indispensable.

### 2. Fine-tuning process

Initially, only the top custom layers were trained, keeping the VGG16 layers frozen. Post this, selective layers of VGG16 were unfrozen, and the model was retrained. This dual-stage training ensured that the model could capitalize on the generic features from ImageNet while also adapting to the nuances of our dataset.

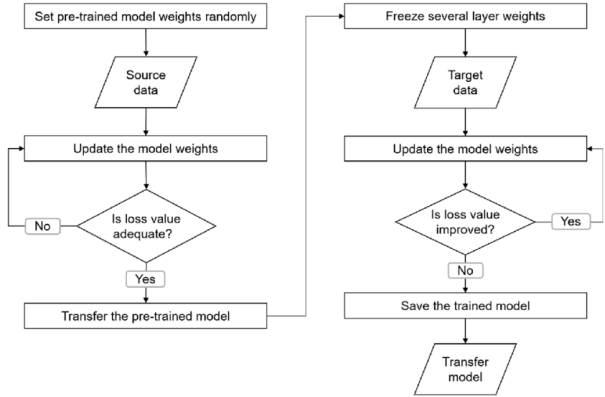

Figure 3.3 Fine-tuning process [24]

## IV. RESULTS AND DISCUSSION

### A. Performance Matrix

The evaluation of our models was grounded on standard metrics - accuracy and loss. These metrics provided a quantitative measure of how well our models were able to classify city images.

Fine-tuned VGG16: Reached a pinnacle with 63.6% accuracy on the Test set.

### B. Comparative Analysis

A comparison between the models underscored the power of transfer learning. The jump in accuracy from the Vanilla CNN to the VGG16 model emphasized the benefits of leveraging pre-trained architectures. Furthermore, the fine-tuning process provided the necessary customization, ensuring the model was tailored to our dataset's specifics.

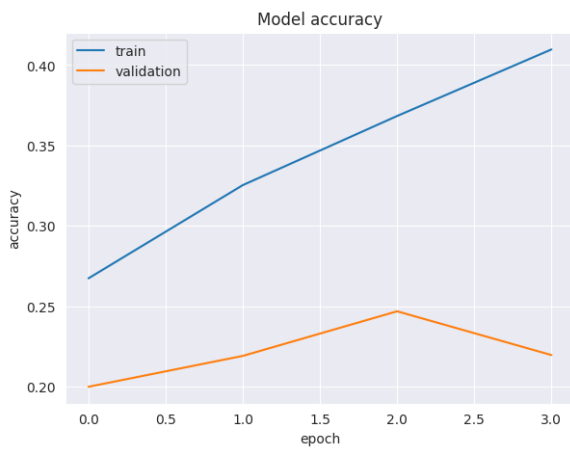

Figure 4.1 : Vanilla CNN (Accuracy)

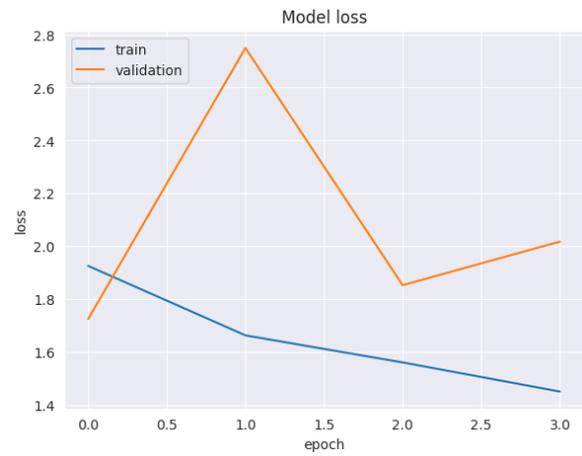

Figure 4.2: Vanilla CNN (Loss)

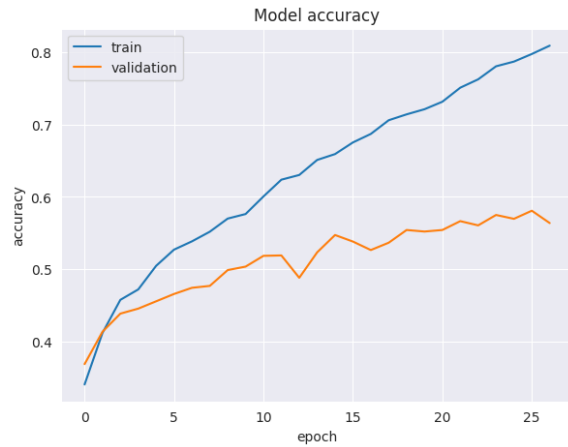

Figure 4.3: VGG16 + Finetune (Accuracy)

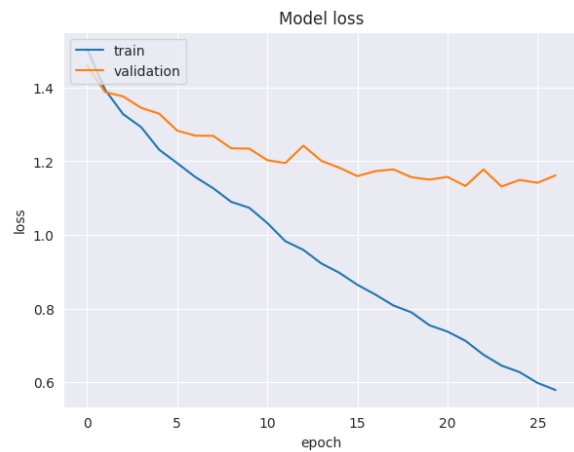

Figure 4.4 VGG16 + finetune (Loss)

### C. Real-world Testing and Observations

On testing our fine-tuned VGG16 model against tools like Google Lens, we observed specific scenarios where our model could recognize city images with higher accuracy. This highlights the value of a specialized tool built and optimized for a niche classification task.

## V. CONCLUSION AND FUTURE SCOPE

### A. Conclusion

Embarking on the journey of creating a specialized image classification system for prominent Indian cities, our research ventured through the realms of deep learning, from constructing a basic Vanilla CNN to harnessing the capabilities of the renowned VGG16 architecture. Through rigorous training, evaluation, and iterative refinement, we realized a model adept at recognizing and distinguishing between the unique visual signatures of various Indian cities.

### B. Future Scope

- Dataset Expansion: Incorporating more cities and diverse images can improve the classifier's robustness.
- Integration with Augmented Reality (AR): AR apps can utilize our classifier to provide real-time city information to users.
- Exploration of other architectures: Beyond VGG16, architectures like ResNet, Inception, and MobileNet might offer enhanced performance or efficiency.
- Real-time feedback loop: An interactive platform where users can correct misclassifications, further training the model in real-time.

## References

[1] Vertongen,p.& Hansen,d., 2008. Location-based Services using Image Search.In: 9th IEEE Workshop on Applications of Computer Vision (WACV 2008), 7-9 January 2008, Copper Mountain, CO, USA.

[2] Jonas Theiner , Eric Muller-Budack , Ralph Ewerth.,2022. Interpretable Semantic Photo Geolocation IEEE/CVF Winter Conference on Applications of Computer Vision (WACV), 2022, pp. 750-760.

[3] A. Feryanto & I. Supriana.,2011. Location recognition using detected objects in an image, 2011 International Conference on Electrical Engineering and Informatics 17-19 July 2011, Bandung, Indonesia.

[4] Viktor B. Shapovalov & Yevhenii B. Shapovalov ,& Zhanna I. Bilyk & Anna P. Megalinskaand & Ivan O. Muzyka.,2019. The Google Lens analyzing quality: an analysis of the possibility to use in the educational process, Publishing center of Kryvyi Rih State Pedagogical University.

[5] Dr. A.Anushya.,2019. Google Lens as an Image Classifier, International Journal of Scientific Research in Computer Science Applications and Management Studies 8(6).

[6] A.Gupta & Dr. Lokesh Kumar.,2021. Andmark recognition using cnn. International Research Journal of Modernization in Engineering Technology and Science.

[7] T.Weyand & I. Kostrikov & J. Philbin.,2016. PlaNet - Photo Geolocation with Convolutional Neural Networks. European Conference on Computer Vision (ECCV) (2016) pp 37–55.

[8] C. Peterson & M. Ludtke., Determining Building Location Based on an Image. University of Wisconsin¬Madison.

[9] Y. Cai & Y. Ding & H. Zhang & J. Xiu & Z. Liu.,2020. Geo-Location Algorithm for Building Targets in Oblique Remote Sensing Images Based on Deep Learning and Height Estimation, Remote Sensing 12(15).

[10] R.Chauhan & K. Kumar Ghanshala & R.C Joshi.,2018. Convolutional Neural Network (CNN) for Image Detection and Recognition, 2018 First International Conference on Secure Cyber Computing and Communication (ICSCCC).

[11] Shuang Li & Baoguo Yu & Yi Jin & Lu Huang .,2021. Image-Based Indoor Localization Using Smartphone Camera. Wireless Communications and Mobile Computing 2021(7):1-9.

[12] Frode Eika Sandnes.,2011. Determining the Geographical Location of Image Scenes based on Object Shadow Lengths, Journal of Signal Processing Systems 65(1):35-47.

[13] I. Maduako , Z. Yi , N. Zurutuza , S. Arora , C. Fabian , D.Hyung Kim.,2022, Automated School Location Mapping at Scale from Satellite Imagery Based on Deep Learning, Remote Sensing 14(4).

[14] Chao-Yeh Chen & K. Grauman.,2011. Clues from the beaten path: Location estimation with bursty sequences of tourist photos, Conference on Computer Vision and Pattern Recognition (CVPR) 20-25 June 2011 ,Colorado Springs, CO, USA.

[15] J. Hays & Alexei A. Efros.,2008. IM2GPS: estimating geographic information from a single image, 2008 IEEE Conference on Computer Vision and Pattern Recognition 23-28 June 2008, Anchorage, AK, USA.

[16] Yukun Yuan.,2022, Geolocation of images taken indoors using convolutional neural network, Monash University, Melbourne, Australia.

[17] N. Samsudin,2021. Photo Geolocation with Neural Networks: How to and How not to available at:https://medium.com/analytics-vidhya/photo-geolocation-with-neural-networks-how-to-and-how-not-to-8aa7f10abb34.

[18] Yunpeng Li & David J. Crandall & Daniel P. Huttenlocher.,2009. Landmark classification in large-scale image collections, 2009 IEEE 12th International Conference on Computer Vision 29 September 2009 - 02 October 2009, Kyoto, Japan

[19] F. Schroff , A. Criminisi , A. Zisserman.,2007. Harvesting image databases from the web, 2007 IEEE 11th International Conference on Computer Vision 4-21 October 2007, Rio de Janeiro, Brazil.

[20] Ghafoorian, M., Karssemeijer, N., Heskes, T. et al. Location Sensitive Deep Convolutional Neural Networks for Segmentation of White Matter Hyperintensities. Sci Rep 7, 5110 (2017). https://www.nature.com/articles/s41598-017-05300-5.

[21] Krizhevsky, A., Sutskever, I., & Hinton, G. E. (2012). ImageNet classification with deep convolutional neural networks. Advances in neural information processing systems, 25

[22] https://www.researchgate.net/publication/340373165/figure/fig1/AS:875683857330176@1585790709597/An-illustration-of-the-architecture-of-the-vanilla-CNN-we-used-for-text-classification.png

[23] https://miro.medium.com/v2/resize:fit:827/1*UeAhoKM0kJfCPA03wt5H0A.png

[24] https://www.researchgate.net/publication/349892484/figure/fig3/AS:999108290551808@1615217387101/Flowchart-of-the-model-transfer-and-the-fine-tuning-procedure.